%% file: root.tex
\documentclass[letterpaper, 10 pt, conference]{ieeeconf}  

\IEEEoverridecommandlockouts                              

\usepackage{graphics} 
\usepackage{epsfig} 
\usepackage{mathptmx} 
\usepackage{times} 
\usepackage{amsmath} 
\usepackage{amssymb}  
\usepackage[]{todonotes}
\usepackage{hyperref}
\hypersetup{hidelinks}

\usepackage{booktabs}
\usepackage{multirow}
\usepackage[table]{xcolor}
\usepackage{siunitx}
\usepackage{pgfplots}
\pgfplotsset{compat=1.18}

\newtheorem{definition}{Definition}
\definecolor{oraclegray}{gray}{0.90}
\usetikzlibrary{backgrounds, calc, positioning,shapes.geometric,fit,shapes.geometric, arrows.meta}

\newcolumntype{G}{>{\columncolor{oraclegray}}c}

\input{commands}

\title{\LARGE \bf
Predictive Zonotope Reduction:\\Precise Runtime Monitoring under Uncertainty
}

\newif\ifpreprint

\preprinttrue      

\ifpreprint
    \author{Vladimir Krsmanovi\' c$^{1*}$, Florian Kohn$^{1}$, Bernd Finkbeiner$^{1,2}$, and Milan Simovi\' c$^{3}$
    \\ \textit{$^{1}$CISPA Helmholtz Center for Information Security \hspace{2pt} $^{2}$Technical University of Munich \hspace{2pt} $^{3}$beanTech}
    \thanks{$^{*}$Correspondence to  {\tt\small vladimir.krsmanovic@cispa.de}}
    }
\else
  \author{Anonymous Authors
  }
\fi

\ifpreprint
    \usepackage[style=ieee]{biblatex}
\else
  \usepackage[style=ieee]{biblatex}
\fi

\begin{document}

\maketitle
\thispagestyle{empty}
\pagestyle{empty}

\begin{abstract}
Robots operating in physical environments make control decisions based on uncertain sensor measurements, which can lead to unsafe or suboptimal actions. Runtime monitors that check their behavior against safety specifications must represent this uncertainty soundly. Zonotopes are a widely used representation, but continuously incorporating new measurements grows their order unboundedly, so monitors must periodically apply an over-approximating reduction. The choice of the reduction method substantially affects the zonotope’s precision, yet existing approaches typically utilize a fixed method throughout the run, even though the optimal choice depends on the current state. This paper presents a Predictive Zonotope Reduction~(PZR) approach, which frames reducer selection as an optimal control problem and solves it using beam-search model predictive control. Policy distillation into a small neural policy further provides substantially higher execution speed than model predictive control while maintaining improved performance, enabling uncertainty-aware runtime monitoring on resource-constrained real-time systems. We implement our approach in the RLola runtime monitoring framework and evaluate it on a 5-degree-of-freedom robotic arm simulated in MuJoCo, with sensor uncertainty modeled according to ISO 5725. Experiments on a Raspberry Pi 5 show that dynamic reduction significantly lowers false-positive rates in monitoring compared with static reduction strategies.

\end{abstract}

\section{Introduction}
\begin{figure*}[ht]
    \centering
    \vspace{3.6pt}
    \includegraphics[width=\linewidth]{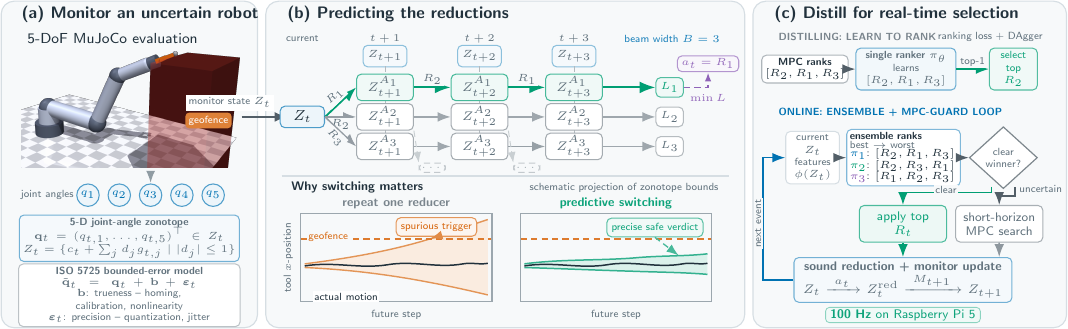}

  \caption{Overview of \textit{Predictive Zonotope Reduction}~(PZR) for uncertainty-aware runtime monitoring.  
  (a) Bounded measurement errors induce a joint-angle zonotope that the monitor checks against a geofence.
  (b) The PZR evaluates sequences of sound zonotope reduction methods and
  applies the first operator from the sequence with the lowest predicted terminal
  approximation loss, which improves accuracy and the false-positive rate
  (c) We distill control policy rankings into lightweight ranking policies.
  At runtime, an ensemble selects a reduction method by voting and consults a control policy in difficult situations. }
\label{fig:overview}
\end{figure*}

Robots operating in the physical world make control decisions based on sensor inputs. Environmental conditions, such as weather, affect the precision with which sensors capture the true physical state of the measured property. A robot that makes control decisions based on uncertain or imprecise sensor inputs can therefore take suboptimal or unsafe control actions.

One solution to ensure the safety of robots operating in uncertain environments is runtime monitoring~\cite{DBLP:conf/cav/rtlolauav,DBLP:conf/sbmf/TeSSla,DBLP:conf/tacas/BiewerFHKSS21}.
In runtime monitoring, a dedicated system component, called a monitor, observes the robot's behavior and validates it against a specification of known good or bad behaviors while the robot is operating. This monitor can then be used to intervene in the event of dangerous robot behavior.
As the monitor bases its assessment of the robot's state on its (uncertain) sensor inputs, it has to track measurement uncertainty to provide sound, precise verdicts.

A common approach for representing interval-bounded measurement uncertainty in such systems is the use of zonotopes~\cite{althoff2021set,Althoff2015ARCH,DBLP:journals/corr/cuttingcorners}. Zonotopes are convex, point-symmetric sets used to efficiently represent the set of system states admissible by the measurements.

However, their use in online settings, that is, while the robot is operating, requires continuously integrating new sensor measurements into the zonotope~\cite{DBLP:conf/rv/rlola,DBLP:journals/automatica/scott}. As more information accumulates in the zonotope, its computational representation grows without bound. This is a major limitation for their application on resource-constrained hardware. In the literature, several zonotope reduction methods address this problem by computing an over-approximation of fixed size~\cite{Althoff2015ARCH,althoff2021set,DBLP:conf/rv/rlola}. In practice, however, the choice of reduction operator is not obvious and is situation-dependent~\cite{kopetzki2017methods}, while strongly affecting the precision of the result.

In this paper, we improve the precision of monitoring verdicts by treating the choice of the reduction method as a sequence problem.
Unlike existing works, which restrict the reduction method to remain the same throughout the system's runtime, we allow different reduction methods to be used over time. As a first solution, we frame this selection process as an optimal control problem, in which the control action at each step is the selection of the optimal zonotope-reduction method. This approach, based on Model Predictive Control~(MPC), improves the false-positive rates of monitoring while additionally removing the need for manual method selection or tuning.

To enable use in real-time systems, we refine the selection method through several optimizations. We enable online operation by introducing causal input prediction and reducing computational overhead through the use of beam search for the control policy. To obtain real-time throughput, we further distill the control policy into a neural policy. To improve robustness to covariate shifts, we combine multiple neural policies with DAgger~\cite{DBLP:journals/jmlr/dagger} augmentations into an ensemble. Lastly, we augment this ensemble with a control policy to handle ambiguous situations and further improve robustness to extreme events while minimally affecting performance. We show that distilling the control policy  into a neural one yields a better policy that runs significantly faster, enabling deployment in real-time systems.

We build our implementation on the RLola monitoring framework~\cite {DBLP:journals/corr/cuttingcorners}, which has seen extensive use in cyber-physical systems~\cite{DBLP:conf/cav/rtlolauav,DBLP:conf/cav/rtloltakeoff,DBLP:conf/tacas/BiewerFHKSS21,DBLP:conf/rv/ros}, and provides formal guarantees about its memory bounds, semantics, and soundness~\cite{DBLP:journals/sttt/rtlolaverguar,DBLP:conf/fm/semantics,DBLP:conf/rv/vermonitors}.
We evaluate our method on a 5-degree-of-freedom robotic arm simulated in MuJoCo~\cite{todorov2012mujoco}, with sensor noise and uncertainty modeled in accordance with ISO~5725~\cite{iso5725-1-2023}. We benchmark our method's performance on a Raspberry Pi~5 equipped with \SI{16}{\giga\byte} of memory.

The results show that our approach outperforms the state of the art by several orders of magnitude. It produces tighter zonotope approximations overall, resulting in a significantly lower false-positive rate (FPR) for the monitor. In practice, a lower FPR means fewer unnecessary emergency stops or task aborts, which is important in scenarios such as robot learning in uncertain environments, where mistakes during exploration can cause harm.
In summary, this paper presents the following contributions:
\begin{itemize} 
\item We formulate \textit{Predictive Zonotope Reduction}~(PZR) as an optimal control problem in which the monitor's action corresponds to choosing an optimal reduction operator. This captures the central observation that zonotope reduction is not merely a local compression step but an operation that affects the future shape of the uncertainty.
\item We instantiate this formulation with a receding-horizon MPC policy that selects reduction methods by optimizing predicted future set tightness. To meet real-time monitoring budgets, we distill the control policy into a DAgger-trained neural ranking policy that selects among sound reduction methods at runtime, composes multiple policies into an ensemble, and integrates the neural and control policies together to improve robustness.
\item We implement predictive reducer selection in the RLola monitoring framework, and evaluate it on robotic systems under realistic measurement uncertainty. Compared with the fixed-reduction-method baselines, our method produces tighter over-approximations, reduces the false-positive rate, and achieves real-time inference speeds. 
\end{itemize}

\section{Background and Related Work}
\subsection{Zonotopes}
A zonotope is a convex, point-symmetric set defined by a center ($c \in \mathbb{R}^d$) and a set of generators ($g_1,\dots,g_k \in \mathbb{R}^d$).
Its points are given by the Minkowski sum of its generators translated by its center:
 \[
\mathbb{Z} \triangleq
\{
c + d_1 g_1 + \cdots + d_k g_k \ \mid\ d_1,\dots,d_k \in [-1,1]
\}.
\]
Zonotopes are widely used in reachability analysis and control theory due to their computational efficiency~\cite{althoff2021set,DBLP:conf/hybrid/Girard05}.
In reachability analysis, zonotopes represent sets of states reachable under all possible input sequences.
In contrast, this paper considers runtime monitoring, which assesses the safety of a single system execution at runtime.
Rather than propagating zonotopes solely according to a system model, runtime monitors continuously refine their state estimates using new sensor measurements.
This is related to set-based observers from control theory~\cite{DBLP:conf/eucc/Combastel03,DBLP:conf/cdc/AlamoB003}, which estimate system states from uncertain observations. The key difference is that runtime monitors aim to provide sound Boolean verdicts about system health rather than precise estimates of an unobservable system state.

All of these applications face the challenge of controlling the order of the zonotope to keep computations feasible. Adaptive reduction methods~\cite{wetzlinger2022adaptive, wetzlinger2020adaptive}, for example, tune their parameters for a given zonotope according to a cost function.
Our approach addresses a different problem: Rather than optimizing the reduction of an individual zonotope, we select reduction methods sequentially across a stream of zonotopes.
This allows the choice of reduction method to depend on the system's evolving state and the anticipated effects of future reductions.

\subsection{Runtime Verification}
Runtime monitoring~\cite{DBLP:journals/jlp/LeuckerS09} is a safety-assurance mechanism widely applied to cyber-physical systems, particularly in safety-critical settings~\cite{DBLP:series/lncs/BartocciDDFMNS18,DBLP:journals/fmsd/MoosbruggerRS17}.
In stream-based runtime monitoring~\cite{DBLP:conf/time/DAngeloSSRFSMM05}, a monitor observes a system through input streams representing sensor measurements.
Output streams define aggregations and transformations of input and other output streams, while triggers specify safety properties that express Boolean verdicts about the system's behavior.

Most stream-based monitoring tools, such as TeSSLa~\cite{DBLP:conf/sbmf/TeSSla}, Striver~\cite{DBLP:conf/rv/Striver}, or Copilot~\cite{DBLP:conf/fm/copilot} do not explicitly account for uncertainty in sensor measurements and may therefore produce verdicts with false confidence. The extension of Lola~\cite{DBLP:conf/time/DAngeloSSRFSMM05} to symbolic inputs~\cite{DBLP:journals/infsof/UnaaLola} models uncertainty through symbolic inputs but relies on interval-based methods, providing overly pessimistic verdicts.

In this paper, we focus on RLola~\cite{DBLP:journals/corr/cuttingcorners}, the robust extension of RTLola~\cite{DBLP:conf/fm/rtlolatutorial} that has seen extensive use in autonomous and safety-critical systems, including UAVs~\cite{DBLP:conf/cav/rtlolauav, DBLP:conf/cav/rtloltakeoff, dlr201362}, as well as deep integration into the robotics software stack via a ROS adapter~\cite{DBLP:conf/rv/ros}.
The tool is validated in practice, supported on different hardware~\cite{DBLP:journals/tecs/fpga,BCFS25}, well documented~\cite{DBLP:conf/fm/rtlolatutorial,DBLP:conf/rv/rtloladesingandintegration}, and provides strong memory guarantees and formally verified compilers and semantics~\cite{DBLP:journals/sttt/rtlolaverguar, DBLP:conf/rv/vermonitors,DBLP:conf/fm/semantics}.

\subsection{Model Predictive Control}
Model Predictive Control (MPC) is a receding-horizon approach to optimal control.
For a system with state $x_k$, control action $u_k$, and dynamics $x_{k+1}=f(x_k,u_k)$, MPC solves the finite-horizon problem
\begin{equation}
  \min_{u_k,\ldots,u_{k+H-1}}
  \sum_{j=0}^{H-1}\ell(x_{k+j},u_{k+j})+V_f(x_{k+H}),
  \label{eq:mpc}
\end{equation}
where $H$ is the horizon, $\ell$ is the stage cost, and $V_f$ is an optional terminal cost. Only the first action of the optimal sequence is applied; after the system advances, the optimization is solved again from the new state.
This feedback loop lets the controller account for delayed consequences of its current action, but it requires a model of how the state evolves under candidate actions and, when relevant, a forecast or preview of exogenous inputs. 

Robust Model Predictive Control is a broad field that focuses on optimizing physical control inputs subject to constraints while accounting for bounded disturbances and model uncertainty~\cite{saltik2018outlook}. While these approaches focus on control \textit{under uncertainty}, we use MPC to control \textit{the evolution of the uncertainty} representation itself - the control actions do not actuate a robot but instead shape the zonotopes.

\section{Predictive Zonotope Reductions}\label{sec:prp}

Existing work treats zonotope reduction as a continuous application of the same method.
This paper views the problem of reducing the order of zonotopes as a sequential decision problem and formulates it as a control problem.
Intuitively, every zonotope reduction affects the future because it shapes the resulting zonotope, which will serve as the basis for future reductions.
Viewing this as a sequential problem, we focus on finding the best reduction methods not just for the current timestep but for the system's entire runtime. 

Varying the zonotope reduction method during the monitor's execution does not affect the approach's validity~\cite{DBLP:journals/corr/cuttingcorners}.
The underlying monitoring algorithm is agnostic of the applied reduction method, and its soundness guarantee for the emitted verdicts relies only on the reduction method chosen at each step to correctly over-approximate the zonotope.
It is then guaranteed that the monitor never misses an actual violation of the specification; that is, false negatives are excluded by construction.
The number of false positives, i.e., the monitor reports a violation even though the specification is not violated, largely depends on the quality of the over-approximation.

We use these insights to formulate zonotope reduction over the full measurement trace as an optimal control problem: We select actions aiming to minimize the final size of the zonotope representing the monitor's state, thereby reducing the uncertainty within it.
The problem of \textit{Predictive Zonotope Reduction}~(PZR) is the optimal control problem of minimizing the loss $\loss$ in the last step of a sequence of zonotope reduction operations.
Such a monitoring sequence is depicted in Fig.~\ref{fig:sequence}.
There, the reference states $\mem_{0}, \dots, \mem_{k}$ represent the most accurate representation of the system state obtainable.
The approximate states $\hmem_{0}, \dots, \hmem_{k}$ represent over-approximations of these states as a result of a sequence of zonotope reduction operations $\redop{b}_0, \dots, \redop{b}_k$.

\input{zigzag}

Framed as a control problem, the system action at each step is the \emph{choice} of a zonotope reduction method from a predefined set. The state of the control problem is the full symbolic monitor state, including its zonotope geometry, and the control action is the selection of a sound reduction method. We use zero-stage cost and the Squared Hull Error relative to an unreduced monitor rollout as the cost, while reducer feasibility and the transform bound provide the constraints. The action is chosen via an MPC loop that optimizes the loss function, which is directly linked to the zonotope's resulting shape. 

To quantify the precision of an over-approximation, we adapt a previously proposed monitoring-specific method~\cite{DBLP:journals/corr/cuttingcorners}. The metric directly estimates the false-positive rate of trigger conditions using zonotopes representing the monitor state. Since trigger conditions correspond to axis-aligned half-space queries over the zonotope, the resulting loss is defined as the dimension-wise mean squared error between the interval hulls of two zonotopes:
\begin{definition}[Squared Hull Error]\label{def:loss}
Let
$Z = (c, g_1, \dots, g_k)$ and
$\hat{Z} = (\hat{c}, \hat{g}_1, \dots, \hat{g}_{\hat{k}})$ be zonotopes in $\mathbb{R}^d$ with $c = \hat{c}$.
The \emph{squared hull error} of $Z$ relative to $\hat{Z}$ is
\[
    \loss(Z, \hat{Z}) \triangleq
    \sum_{i=1}^{d}
        \Bigg(
            \sum_{j}^{k} |g_{j,i}| \;-\; \sum_{j}^{\hat{k}} |\hat{g}_{j,i}|
        \Bigg)^{2}
\]
\end{definition}

At each time step, Predictive Zonotope Reductions selects the next reduction action by minimizing predicted Squared Hull Error over a future finite sequence of decisions. For a decision depth $H$ that counts the current event, let $Z_{t+H}^{A}$ be the zonotope state obtained by rolling out a sequence of zonotope reduction actions $A=(a_t,\ldots,a_{t+H-1})$ from a set of all possible method combinations $A^H$ over the fixed finite horizon $H$. Let $Z_{t+H}$ be the reference zonotope on which no reduction method was applied. We pick the next zonotope reduction method $a_t$ by finding the optimal sequence of reductions $A^{\star}$ over the horizon, and select the starting action of that sequence as $a_t = (A^\star)_t$.
\begin{equation}
   A^\star\in\arg\min_{A \in A^{H}}
  \loss \bigl(Z_{t+H}^{A},Z_{t+H}\bigr),
  \label{eq:predictive-reduction}
\end{equation}
We perform this optimization search in every state of the system in which the zonotope size exceeds the bound $b$, with no reduction being applied if the size is within the bound. The search discards infeasible candidates when the method fails due to broken assumptions and ensures that a feasible selection method always exists. The selected action $a_t$ gets applied, and the search is repeated in the next timestep.

\subsection{Recorded-Future Reductions}
In offline monitoring, where the traces are prerecorded, or where the zonotope-reduction decision is staggered (effectively giving us a Perfect Event Preview), the future events are precisely known. Importantly, known measurements of the system do not eliminate uncertainty, as each recorded value remains subject to bounded measurement error within the monitor. In this case, the control decision on which zonotope reduction method to select can be made precisely based on the known future inputs. Starting from the current monitor state, we roll each candidate schedule through the recorded current and future events, with an unreduced branch (with no zonotope reductions applied) serving as the reference. We refer to this method, which uses a full rollout with the known three future inputs, as \textit{MPC-F}.

However, enumerating all possible combinations of method selections over the horizon is computationally prohibitive due to the exponential size of the space. To address this and enable the use of more methods and a longer horizon, we adapt beam search~\cite{lowerre1976harpy} as a constrained search method. After each event, the search retains the $W$ prefixes with the lowest current terminal loss, expands them with every reducer, and finally commits the first action of the best complete schedule. We refer to this variant as \textit{MPC-B}, which considers the next five future input events and explores them using a beam search with width $W=4$.

\subsection{Causal Predictive Reductions}
For a more realistic scenario, we have to consider an online setting in which future events are not precisely known and must be predicted. We adapt our approach to this setting by implementing a causal prediction step that predicts future events for the MPC rollout from previously observed inputs. While these predictions can lead to different reducer choices and cannot predict outlier events, they provide a strong basis for the online setting of our method without sacrificing soundness and safety, as these guarantees are preserved by our zonotope reduction methods. The method \textit{MPC-L} performs Predictive Zonotope Reductions with a Horizon Length of $H=5$ and a beam search width of $W=4$. Future inputs are predicted by linearly extrapolating from previously seen inputs.

\subsection{Learned Policy Distillation}\label{subsec:distil}
While the causal prediction of future inputs enables our method to be used in the online setting, where these inputs are not known in advance, the computational bottleneck of MPC remains restrictive for realistic deployments. To enable this, we \textit{distill} the control policies into a small neural network. The neural policies act as a student network, learning from the teacher outputs of the control policies, as determined by one of the previously mentioned Predictive Zonotope Reduction methods. 

We set up the teacher-student setting with the same possible reduction actions, but focus the neural policy on minimal resource requirements and maximal decision throughput. While the control policies can roll out future inputs and search for the optimal sequence of reductions, we limit the student inputs to scenario-agnostic features of the current zonotope. While this leads to an information-asymmetric setting, which can limit performance, several adaptations preserve our approach's performance.

Crucially, the student policy is not trained on a classification objective (``what is the optimal reduction method for this step'') but on a current-state ranking policy. In each state, the student policy must provide a pairwise ranking of possible reducers that the network can choose. These pairwise rankings yield a ranked list of preferred reduction methods for each state. This learning objective provides a significantly denser learning signal, especially given the imbalance in method-selection choices produced by the control policies (as discussed and shown in Sec.~\ref{sec:evaluation}). Because the student network can choose only among sound reduction methods, it preserves the same soundness guarantees as the teacher-control policies.  

As the learned policies are trained via distillation on sampled traces, the resulting policies are vulnerable to the well-known issue of covariate shift.  To improve their robustness, we combine several learned policies into an ensemble. At each step, the policies independently rank the reduction operators and vote
for the operator to apply, with ties resolved using rank and confidence.  To improve overall robustness and introduce diversity into the ensemble, each policy is independently augmented with sampled traces using DAgger~\cite{DBLP:journals/jmlr/dagger}. We refer to these variants as \textit{Vote}. To further improve robustness and reduce the covariate shift, we extend the ensemble with a control policy \textit{Vote-Guarded}. In uncertain cases, where the ensemble votes do not select a clear winner, rather than breaking ties based on secondary metrics, the next reduction method is selected based on a prediction made by one of the control policies using a short horizon and beam width.

\section{Evaluation}
\label{sec:evaluation}
We evaluate our methods on the robot-arm monitoring scenario illustrated in Fig.~\ref{fig:overview}. We simulate a 5-DoF MuJoCo robot arm~\cite{todorov2012mujoco} at \SI{10}{\hertz} and model measurement error according to  ISO~5725~\cite{iso5725-1-2023}.
The error model includes quantization error, jitter, integral nonlinearity due to magnet displacement, and homing and calibration errors. The robot arm is monitored by an RLola monitor against a specification that prevents it from entering a forbidden region. We run the experiments on a Raspberry Pi 5 with \SI{16}{\giga\byte} of memory running Raspbian 64-bit.

\subsection{Experimental Setup}

\paragraph{Methods}
Predictive Zonotope Reduction policies choose from four reduction methods described in existing literature - namely Girard~\cite{DBLP:conf/hybrid/Girard05}, Scott~\cite{DBLP:journals/automatica/scott}, PCA~\cite{kopetzki2017methods}, and Combastel~\cite{DBLP:conf/eucc/Combastel03} reducers. While many other methods exist, experiments have shown that they are rarely selected by policies and that their standalone performance is significantly worse than that of the four mentioned methods. We implement these methods in RLola, with our Rust implementations tracking their implementation in the CORA tool~\cite{Althoff2015ARCH}.

\paragraph{Controllers}  
For control policies, we evaluate on the previously mentioned variations of MPC policies. MPC-B and MPC-F have perfect previews of future inputs, while MPC-L has to predict them causally. The policies are evaluated over the horizon length of $H=5$, and MPC-L and MPC-B search over the beam width of $W=4$. 
We introduce the MPC-Greedy ($H=1$) baseline, which selects the best next action at the current state and can be viewed as a restricted version of our methods, highlighting the importance of switching between zonotope reduction methods. 
 
We construct the neural policies as two dense layers of 32 neurons each, with ReLU activation, mapping the 15 input zonotope properties to 4 lower-is-better preference scores, thereby creating a ranking of methods. As discussed in Sec.~\ref{subsec:distil}, the policies do not have access to information such as stream values, history, or predicted rollouts. They are distilled over 128 randomly sampled traces using MPC-F as the teacher policy. We refer to this neurally distilled policy as G15. For the ensemble method described before, we combine three G15 policies, each augmented with 6 additional independently sampled random traces. The Vote3Guard method follows the control policy guard in case the three policies do not unanimously select a reduction method.

\paragraph{Cohorts and Metrics} 
We evaluate the performance of our methods across a range of sampled robot arm behaviors and zonotope budget bounds (the maximal shape of the zonotope representation), showcasing a connection between method selection and memory bounds $b\in\{40,80,120,150,200,250,500\}$. We generate 20 random-waypoint executions and monitor the traces at \SI{10}{\hertz} using the geofence specification.

Our primary metric is the Squared Hull Error $\loss$ (from Definition~\ref{def:loss}), computed with respect to a ground-truth reference zonotope. We also report the false-positive rate (FPR), whose denominator is the number of negative reference events, and the p99 latency for reduction method selection and execution. While $\loss$ and the FPR are correlated, they need not coincide, as a larger zonotope can still be better aligned with the monitor predicate. We note that the false-negative rate is 0 across all methods, and we do not report it in tables or figures for conciseness. All controllers measured use sound overapproximation methods; thus, reductions can affect tightness and false interventions but not soundness\footnote{Nevertheless, we track the metric during evaluations and confirm that no evaluated trace produces a false negative over approximately 700{,}000 monitored events }.

\subsection{Results and Analysis}
\begin{figure*}[ht]
  \centering
  \includegraphics[width=0.97\linewidth]{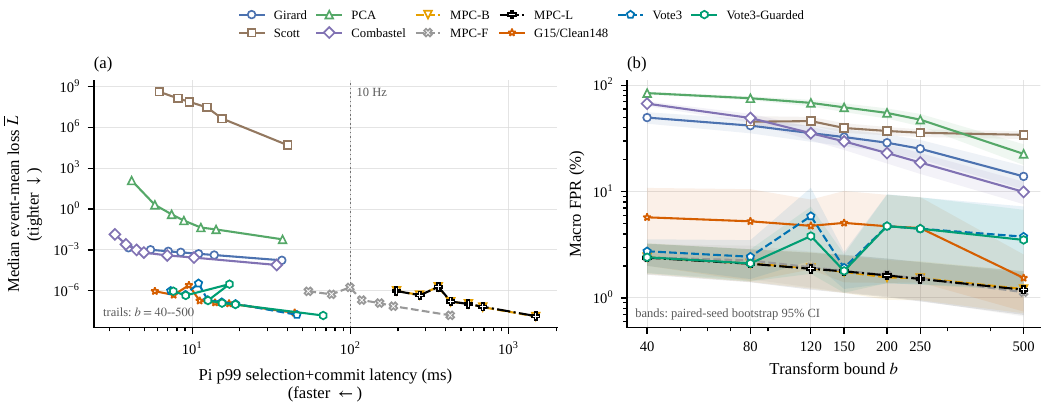}
  \caption{
    Quality--latency trade-off across transform bounds; lower is better on every vertical axis.
    (a) Median event-mean approximation loss over 20 traces against median p99 selection-plus-commit latency over five Raspberry Pi~5 traces.
    (b) FPR over 20 traces with paired-seed bootstrap \SI{95}{\percent} intervals.
  }
  \label{fig:tradeoff}
\end{figure*}
In total, we performed 1,400 monitoring runs, and present the results and their analysis in the following section.

\begin{figure}[ht]
  \centering
  \includegraphics[width=\columnwidth]{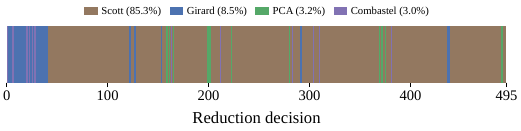}
  \caption{Method selections of MPC-L on a selected trace showing the ration and distribution of reduction method selection decisions}
  \label{fig:reducer-timeline}
\end{figure}

\begin{figure}[ht]
  \centering
  \includegraphics[width=\columnwidth]{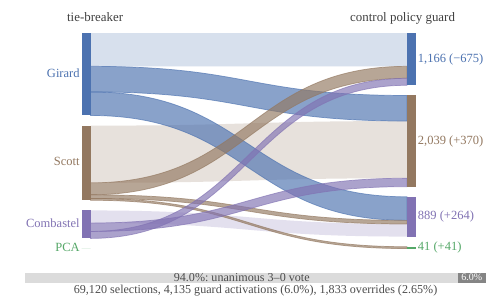}
  \caption{
    Changes in selected reduction methods between tie-breaking and a control policy guard over 69{,}120 selections.
    Pale bands preserve the vote winner, saturated bands are overrides, and brackets indicate net changes.
  }
  \label{fig:guard}
\end{figure}

\begin{figure}[th]
  \centering
  \vspace{3.3pt}
  \includegraphics[width=\columnwidth]{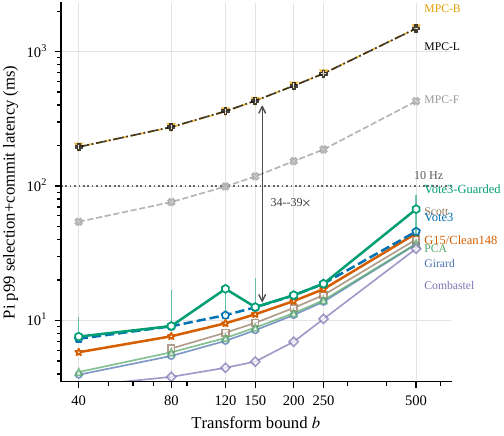}
  \caption{
    Pi p99 latency across bounds.
    Points are five-seed medians, whiskers are observed ranges, and the dotted line marks \SI{100}{\milli\second}.
  }
  \label{fig:pi-sweep}
\end{figure}

Table~\ref{tab:headline} summarizes the performance of all methods at $b=150$, while Fig.~\ref{fig:tradeoff} reports approximation loss, FPR, and selection-plus-commit latency across all transform bounds.   Table~\ref{tab:reducer-composition} and  Fig.~\ref{fig:reducer-timeline} summarize reducer choices,   Fig.~\ref{fig:guard} shows the behaviour of the control policy guard, and Fig.~\ref{fig:pi-sweep} reports Raspberry Pi~5 latency across budget bounds.

\paragraph{\textbf{Dynamic Selection beats Fixed Reduction Methods}}
\label{sec:eval-fixed}
Table~\ref{tab:headline} shows that Predictive Zonotope Reductions produce tighter zonotopes and fewer false positives than every fixed reducer.   At $b=150$, the best fixed reducer has an FPR of \SI{29.68}{\percent}, compared with  \SIrange{1.76}{5.08}{\percent} for the dynamic selectors. Fig.~\ref{fig:tradeoff} shows that this advantage persists across the transform bounds. Furthermore, the performance and throughput plot in Fig.~\ref{fig:tradeoff} shows that fixed schedules are fast but loose, that PZR control policies are precise but computationally expensive, and the neural PZR policies retain MPC-scale loss while achieving the speed required for real deployment.

\begin{table}[ht]
  \centering
  \caption{Median $\loss$ and FPR over 20 held-out traces  at $b=150$.}
  \label{tab:headline}
  \small
  \setlength{\tabcolsep}{3.5pt}
  \input{generated/headline_results.tex}
\end{table}
MPC-Greedy ($H=1$) is the simplest PZR instance: it exhaustively scores the four reducers for the current decision without a future rollout. At $b=150$, switching alone reduces the best fixed median loss from \num{6.4e-04} to \num{2.6e-07} and the FPR from \SI{29.68}{\percent} to \SI{1.96}{\percent}. MPC-L then scores four causally predicted future events in addition to the current decision, reaching \num{1.6e-07} loss and \SI{1.78}{\percent} FPR; across the full paired sweep, it improves over Greedy in 107 of 140 cells.

\paragraph{\textbf{Reduction Methods act complementary over long traces}}
The analysis of the selected methods shows that the performance of reduction methods does not strictly correlate with their performance when combined alongside other methods over the full execution of the system. Table~\ref{tab:reducer-composition} shows the frequency with which the PZR policies select specific reduction methods.

\begin{table}[ht]
  \centering
  \caption{Reducer-selection frequencies (\%) of different policies. }
  \label{tab:reducer-composition}
  \small
  \input{generated/reducer_composition.tex}
\end{table}

Perhaps surprisingly, the dynamic controllers choose Scott most often of all methods despite its poor standalone performance. We attribute this to the Scott method being vulnerable to poorly shaped matrices, which often lead to poor performance or even complete failure\footnote{Scott's method assumes the zonotope matrix is invertible, which often fails during longer executions}. However, the application of the Girard or PCA methods at the beginning of the trace, and their occasional use throughout (as shown in Fig.~\ref{fig:reducer-timeline}), is crucial. We hypothesize that their use conditions the zonotope representation into a form optimal for applying Scott's methods, leading to significant performance gains observed in Fig.~\ref{fig:tradeoff}. This shows the contrast between the dynamic and static zonotope reduction settings, as well as the complementary interactions among the various methods.

\paragraph{\textbf{Learned Policies can learn the control behavior with minimal inputs}}
The results in Tables~\ref{tab:headline} and Fig.~\ref{fig:tradeoff} show that learned policies distilled from the control policies can retain strong performance despite having access to only a limited set of information compared to the control policies, and that they significantly outperform the static methods at minimal runtime cost. 

However, a deeper analysis of the method's performance reveals that simple distillation occasionally fails under covariate shift. These occasional failures motivate the ensemble and control policy guard measures introduced in  Sec.~\ref{subsec:distil}. As shown in Table~\ref{tab:headline} and Fig.~\ref{fig:tradeoff}, these additions help mitigate these issues, although they do not completely resolve them. We find that Vote3-Guarded helps recover performance on the problematic traces compared to the plain learned policies.

Table~\ref{tab:reducer-composition} and Fig.~\ref{fig:guard} show how the control policy guard helps resolve part of these issues. An interesting analysis emerges from examining the Guard method's actions, which override those of the ensembles when the ensembles fail to select a method unanimously. We find it turns on rarely and does not always strictly replace the method that would have been selected via tie-breaking, as in the pure ensemble variations. Looking at  69{,}120 method selection decisions, \SI{94.02}{\percent} were unanimous, while the guard was activated on the remaining \SI{5.98}{\percent}, of which \SI{2.65}{\percent} led to a method change compared to the tie-breaker method.

As shown in Fig.~\ref{fig:guard}, most of the overrides seem to encourage exploitation of the Scott methods' performance, with occasional redirection from Scott to Girard or PCA. These actions help stabilize the policy behavior by encouraging active exploitation of Scott when available, while also helping to avoid failure modes by occasionally intervening with PCA and Girard overrides to condition the matrices for follow-on reductions.

While the occasional failures of the learned policies are a limitation of the method, several ways of fixing it remain open for future work, such as shadow backup copies, the addition of history, or, most directly, removing the information asymmetry by adopting a more powerful input policy. Importantly, this limitation never threatens the system's safety, as the construction of the Predictive Zonotope Reductions ensures safety-by-construction.

\paragraph{\textbf{Learned Polies Can Run Fast}}
\label{sec:eval-pi}
We demonstrate that our method can perform runtime monitoring of embedded systems, with minimal resources and high throughput. We profile the latency of our methods on a single isolated Cortex-A76 core of a Raspberry Pi~5. Figs.~\ref{fig:pi-sweep}~and~\ref{fig:tradeoff} show that the learned selectors combine low approximation loss of the control policies with low latency. Control policies can sustain the 10Hz throughput at its lower bound, but, more importantly, we show that the learned policies enable a significant boost in throughput while maintaining high accuracy as can be observed in Fig.~\ref{fig:tradeoff}.

This also applies to the Vote3 Guarded method, which occasionally runs a control policy, but its selective use still keeps its performance close to that of static methods such as Scott or Girard. Overall, neural PZR policies distilled from the control PZR can combine the performance and high throughput required for deployments. The distilled neural policies introduce negligible overhead and are primarily bottlenecked by the throughput of the static reduction methods they select. This shows potential for even faster performance on hardware with greater processing power or multi-core capabilities.

\section{Limitations}
Our policies select among existing zonotope-reduction methods (we do not construct new ones), and our method's performance is tied to theirs. Our work shows that switching between methods, as well as predicting future inputs, provides significant benefits, and we view it as deeply complementary. This, however, comes at a computational cost, and the balance among horizon, beam width, and size bounds depends on the specific problem and deployment. As the physical ground truth is generally unknown, we compute the loss metrics against a predicted ground truth derived online from unreduced rollouts - this is an unavoidable part of working with uncertainty. The learned policies presented aim to maximize throughput on resource-constrained embedded hardware, which is prone to covariate shift. Importantly, these issues never risk loss of soundness or safety - they only affect the zonotope tightness.

\section{Conclusion}
In this paper, we introduced Predictive Zonotope Reduction, a sequential control framework for online uncertainty monitoring in autonomous systems. By treating zonotope reduction as an optimal control problem rather than applying a static method throughout the run, our approach significantly improves representation accuracy by orders of magnitude and reduces vacuous alarms arising from false positives. 

To satisfy embedded real-time constraints, we distilled receding-horizon MPC controllers into lightweight neural ranking policies augmented with a DAgger ensemble and a short-horizon control guard. We evaluated our method on a 5-DoF robotic arm simulated in MuJoCo, with measurement uncertainty per ISO 5725. Deployed on a Raspberry Pi 5, we sustain high-frequency monitoring while preserving formal safety guarantees by construction. 
\ifpreprint

    \section{Acknowledgments} This work was partially supported by the German Research Foundation (DFG) as part of PreCePT (FI 936/7-1; FR 2715/6-1) and partially supported by the European Research Council (ERC) Grant HYPER (No.~101055412). Views and opinions expressed are, however, those of the author(s) only and do not necessarily reflect those of the European Union or the European Research Council Executive Agency. Neither the European Union nor the granting authority can be held responsible for them. Florian Kohn carried out this work as a member of the Saarbrücken Graduate School of Computer Science.
\fi

\printbibliography[]

\end{document}

%% file: commands.tex
\newcommand{\loss}{\overline{L}}
\newcommand{\mem}{Z}
\newcommand{\hmem}{{Z^A}}
\newcommand{\redop}[1]{a}


%% file: zigzag.tex

\begin{figure}[ht]
\centering
\resizebox{\columnwidth}{!}{%
\begin{tikzpicture}[
    x=1cm,
    y=1cm,
    reference/.style={
        draw=black!75,
        fill=blue!8,
        rounded corners=1.5pt,
        font=\footnotesize,
        inner sep=2pt
    },
    bounded/.style={
        draw=black!75,
        fill=orange!15,
        rounded corners=1.5pt,
        font=\footnotesize,
        minimum width=0.62cm,
        minimum height=0.48cm,
        inner sep=2pt
    },
    expanded/.style={
        draw=black!65,
        fill=black!5,
        rounded corners=1.5pt,
        font=\footnotesize,
        minimum width=0.84cm,
        minimum height=0.56cm,
        inner sep=2pt
    },
    update/.style={
        -{Stealth[length=1.5mm]},
        semithick,
        black!70
    },
    reduction/.style={
        -{Stealth[length=1.5mm]},
        semithick,
        densely dashed,
        black!70
    },
    loss/.style={
        {Stealth[length=1.2mm]}-{Stealth[length=1.2mm]},
        densely dotted,
        black!45
    },
    transition label/.style={
        font=\footnotesize,
        inner sep=1pt
    },
    lane label/.style={
        font=\footnotesize\itshape,
        black!75,
        align=left
    },
    annotation/.style={
        font=\scriptsize,
        black!45,
        align=left
    }
]

\path[use as bounding box]
    (-1.48,0.55) rectangle (7.40,-3.52);

\def\xzero{0}
\def\xone{2.05}
\def\xtwo{4.10}
\def\xthree{6.15}

\def\referenceY{0}
\def\boundedY{-1.45}
\def\expandedY{-2.55}


\node[
    lane label,
    anchor=west
] at (-1.43,-0.62) {Full-order\\reference};

\node[
    reference,
    minimum width=0.60cm,
    minimum height=0.44cm
] (z0) at (\xzero,\referenceY) {$Z_0$};

\node[
    reference,
    minimum width=0.80cm,
    minimum height=0.52cm
] (z1) at (\xone,\referenceY) {$Z_1$};

\node[
    reference,
    minimum width=1.08cm,
    minimum height=0.64cm
] (z2) at (\xtwo,\referenceY) {$Z_2$};

\node[
    reference,
    minimum width=1.46cm,
    minimum height=0.82cm
] (z3) at (\xthree,\referenceY) {$Z_3$};

\node[
    font=\footnotesize
] (reference-dots) at (7.25,\referenceY) {$\cdots$};

\draw[update]
    (z0.east) --
    node[
        transition label,
        above=2pt
    ] {$M_0$}
    (z1.west);

\draw[update]
    (z1.east) --
    node[
        transition label,
        above=2pt
    ] {$M_1$}
    (z2.west);

\draw[update]
    (z2.east) --
    node[
        transition label,
        above=2pt
    ] {$M_2$}
    (z3.west);

\draw[update]
    (z3.east) --
    (7.07,\referenceY);


\node[
    lane label,
    anchor=west
] at (-1.43,\boundedY-0.61) {Bounded\\execution};

\node[bounded] (zh0)
    at (\xzero,\boundedY) {$\widehat Z_0$};

\node[bounded] (zh1)
    at (\xone,\boundedY) {$\widehat Z_1$};

\node[bounded] (zh2)
    at (\xtwo,\boundedY) {$\widehat Z_2$};

\node[bounded] (zh3)
    at (\xthree,\boundedY) {$\widehat Z_3$};

\node[expanded] (zp1)
    at ({(\xzero+\xone)/2},\expandedY)
    {$\widehat Z'_1$};

\node[expanded] (zp2)
    at ({(\xone+\xtwo)/2},\expandedY)
    {$\widehat Z'_2$};

\node[expanded] (zp3)
    at ({(\xtwo+\xthree)/2},\expandedY)
    {$\widehat Z'_3$};

\node[
    font=\footnotesize
] (bounded-dots) at (7.25,\expandedY) {$\cdots$};

\draw[update]
    (zh0.south east) --
    node[
        transition label,
        pos=0.48,
        left=2pt
    ] {$M_0$}
    (zp1.north west);

\draw[reduction]
    (zp1.north east) --
    node[
        transition label,
        pos=0.48,
        left=2pt
    ] {$R_b^{a_0}$}
    (zh1.south west);

\draw[update]
    (zh1.south east) --
    node[
        transition label,
        pos=0.48,
        left=2pt
    ] {$M_1$}
    (zp2.north west);

\draw[reduction]
    (zp2.north east) --
    node[
        transition label,
        pos=0.48,
        left=2pt
    ] {$R_b^{a_1}$}
    (zh2.south west);

\draw[update]
    (zh2.south east) --
    node[
        transition label,
        pos=0.48,
        left=2pt
    ] {$M_2$}
    (zp3.north west);

\draw[reduction]
    (zp3.north east) --
    node[
        transition label,
        pos=0.48,
        left=2pt
    ] {$R_b^{a_2}$}
    (zh3.south west);

\draw[update]
    (zh3.south east) --
    node[
        transition label,
        pos=0.48,
        left=2pt
    ] {$M_3$}
    (7.05,\expandedY);


\draw[loss]
    (z0.south) --
    node[
        transition label,
        right=2pt,
        fill=white,
        inner sep=0.5pt
    ] {$\widehat L$}
    (zh0.north);

\draw[loss]
    (z1.south) --
    node[
        transition label,
        right=2pt,
        fill=white,
        inner sep=0.5pt
    ] {$\widehat L$}
    (zh1.north);

\draw[loss]
    (z2.south) --
    node[
        transition label,
        right=2pt,
        fill=white,
        inner sep=0.5pt
    ] {$\widehat L$}
    (zh2.north);

\draw[loss]
    (z3.south) --
    node[
        transition label,
        right=2pt,
        fill=white,
        inner sep=0.5pt
    ] {$\widehat L$}
    (zh3.north);


\coordinate (key-left) at (-0.25,-3.34);
\coordinate (key-right) at (3.50,-3.34);

\draw[update]
    (key-left) -- ++(0.48,0);

\node[
    annotation,
    anchor=west
] at ($(key-left)+(0.64,0)$)
    {monitoring adds generators};

\draw[reduction]
    (key-right) -- ++(0.48,0);

\node[
    annotation,
    anchor=west
] at ($(key-right)+(0.64,0)$)
    {reduction restores the bound};

\end{tikzpicture}%
}
\caption{The sequence problem visualized. $M_0,\dots,M_k$ are monitoring steps that transform the state zonotopes $\mem_i$ and $\hmem_i$ given the input measurements. $\redop{b}_0,\dots,\redop{b}_k$ is the sequence of zonotope reduction operators.}
\label{fig:sequence}
\end{figure}

%% file: generated/headline_results.tex
\begin{tabular}{l r S[table-format=2.2]}
\toprule
Method & {$\overline{L}\downarrow$} & {FPR (\%)\,$\downarrow$} \\
\midrule
\multicolumn{3}{l}{\emph{Fixed reducer}} \\
\quad Girard & \num{6.7e-04} & 32.60 \\
\quad Scott & \num{7.6e+07} & 39.70 \\
\quad PCA & \num{1.4e-01} & 62.11 \\
\quad Combastel & \num{6.4e-04} & 29.68 \\
\multicolumn{3}{l}{\emph{Recorded-Future Reductions}} \\
\quad MPC-F & \num{2.1e-07} & 1.80 \\
\quad MPC-B & \num{1.5e-07} & 1.76 \\
\multicolumn{3}{l}{\emph{Causal Predictive Reductions}} \\
\quad MPC-Greedy ($H=1$) & \num{2.6e-07} & 1.96 \\
\quad MPC-L & \bfseries \num{1.6e-07} & \bfseries 1.78 \\
\quad G15/Clean148 & \num{1.9e-07} & 5.08 \\
\quad Vote3 & \num{2.1e-07} & 1.92 \\
\quad Vote3-Guarded & \num{1.9e-07} & 1.79 \\
\bottomrule
\end{tabular}

%% file: generated/reducer_composition.tex
\begin{tabular}{l *{4}{S[table-format=2.2]}}
\toprule
Controller & {Girard} & {Scott} & {PCA} & {Combastel} \\
\midrule
\quad MPC-B & 6.73 & \bfseries 85.55 & 3.03 & 4.69 \\
\quad MPC-F & 8.65 & \bfseries 84.24 & 2.38 & 4.73 \\
\quad MPC-L & 6.72 & \bfseries 85.56 & 3.20 & 4.52 \\
\quad G15/Clean148 & 10.99 & \bfseries 88.00 & 0.46 & 0.56 \\
\quad Vote3 & 10.41 & \bfseries 87.69 & 0.25 & 1.66 \\
\quad Vote3-Guarded & 9.13 & \bfseries 89.12 & 0.06 & 1.69 \\
\bottomrule
\end{tabular}

%% file: ref.bib
@article{althoff2021set,
  title={Set propagation techniques for reachability analysis},
  author={Althoff, Matthias and Frehse, Goran and Girard, Antoine},
  journal={Annu. Rev. Control Robot. Auton. Syst.},
  year={2021},
}

@inproceedings{DBLP:conf/rv/ros,
  author       = {Jan Baumeister and
                  Bernd Finkbeiner and
                  Franz J{\"{u}}nger and
                  Florian Kohn and
                  Sebastian Schirmer and
                  Christoph Torens},
  title        = {A {ROS} Adapter for RTLola},
  booktitle    = {RV 2025},

}

@article{DBLP:journals/corr/cuttingcorners,
  author       = {Bernd Finkbeiner and
                  Martin Fr{\"{a}}nzle and
                  Florian Kohn and
                  Paul Kr{\"{o}}ger},
  title        = {Cutting Corners on Uncertainty: Zonotope Abstractions for Stream-based
                  Runtime Monitoring},
  year         = {2026}
}

@inproceedings{Althoff2015ARCH,
	author			= {Matthias Althoff},
	title			= {An Introduction to {CORA} 2015},
	booktitle		= {ARCH},
	year			= {2015},
  }

@inproceedings{DBLP:conf/fm/rtlolatutorial,
  author       = {Jan Baumeister and
                  Bernd Finkbeiner and
                  Florian Kohn and
                  Frederik Scheerer},
  title        = {A Tutorial on Stream-Based Monitoring},
  booktitle    = { {FM} 2024},
}

@inproceedings{DBLP:conf/cav/rtlolauav,
  author       = {Jan Baumeister and
                  Bernd Finkbeiner and
                  Florian Kohn and
                  Florian L{\"{o}}hr and
                  Guido Manfredi and
                  Sebastian Schirmer and
                  Christoph Torens},
  title        = {Monitoring Unmanned Aircraft: Specification, Integration, and Lessons-Learned},
  booktitle    = {{CAV} 2024},
}

@article{DBLP:journals/sttt/rtlolaverguar,
  author       = {Jan Baumeister and
                  Johann C. Dauer and
                  Bernd Finkbeiner and
                  Sebastian Schirmer},
  title        = {Monitoring with verified guarantees},
  journal      = {Int. J. Softw. Tools Technol. Transf.},
  year         = {2023},
}

@inproceedings{DBLP:conf/cav/rtloltakeoff,
  author       = {Jan Baumeister and
                  Bernd Finkbeiner and
                  Sebastian Schirmer and
                  Maximilian Schwenger and
                  Christoph Torens},
  title        = {RTLola Cleared for Take-Off: Monitoring Autonomous Aircraft},
  booktitle    = { {CAV}  2020},
}

@inproceedings{DBLP:conf/rv/rtloladesingandintegration,
  author       = {Maximilian Schwenger},
  title        = {Monitoring Cyber-Physical Systems: From Design to Integration},
  booktitle    = {{RV} 2020},

}

@article{DBLP:journals/tecs/fpga,
  author       = {Jan Baumeister and
                  Bernd Finkbeiner and
                  Maximilian Schwenger and
                  Hazem Torfah},
  title        = {{FPGA} Stream-Monitoring of Real-time Properties},
  journal      = {{ACM} Trans. Embed. Comput. Syst.},
  year         = {2019},
}

@inproceedings{DBLP:conf/rv/vermonitors,
  author       = {Bernd Finkbeiner and
                  Stefan Oswald and
                  Noemi Passing and
                  Maximilian Schwenger},
  title        = {Verified Rust Monitors for Lola Specifications},
  booktitle    = {{RV} 2020},
}

@inproceedings{DBLP:conf/fm/semantics,
  author       = {Florian Kohn and
                  Arthur Correnson and
                  Jan Baumeister and
                  Bernd Finkbeiner},
  title        = {Pacing Types for Asynchronous Stream Equations},
  booktitle    = {{FM} 2026},
}

@inproceedings{dlr201362,
            year = {2024},
           title = {Certification Aspects of Runtime Assurance for Urban Air Mobility},
          author = {Torens, Christoph and Nagarajan, Pranav and Schirmer, Sebastian and Dauer, Johann C. and Baumeister, Jan and Kohn, Florian and Finkbeiner, Bernd and L{\"o}hr, Florian and Manfredi, Guido},
       publisher = {AIAA},
}

@inproceedings{BCFS25,
  author    = {Jan Baumeister and
			   Arthur Correnson and
               Bernd Finkbeiner and
               Frederik Scheerer},
  title        = {An Intermediate Program Representation for Optimizing Stream-Based Languages},
  booktitle    = {CAV 2025},
}

@inproceedings{todorov2012mujoco,
  title={MuJoCo: A physics engine for model-based control},
  author={Todorov, Emanuel and Erez, Tom and Tassa, Yuval},
  booktitle={2012 IEEE/RSJ},

}

@techreport{iso5725-1-2023,
  author      = {{International Organization for Standardization}},
  title       = {Accuracy (trueness and precision) of measurement methods and results --- Part 1: General principles and definitions},
  number      = {ISO 5725-1:2023},
  type        = {Standard},
  edition     = {2},
}

@inproceedings{DBLP:conf/tacas/BiewerFHKSS21,
  author       = {Sebastian Biewer and
                  Bernd Finkbeiner and
                  Holger Hermanns and
                  Maximilian A. K{\"{o}}hl and
                  Yannik Schnitzer and
                  Maximilian Schwenger},

  title        = {RTLola on Board: Testing Real Driving Emissions on your Phone},
  booktitle    = {{TACAS} 2021},
  year         = {2021},
}

@inproceedings{DBLP:journals/jmlr/dagger,
  author       = {St{\'{e}}phane Ross and
                  Geoffrey J. Gordon and
                  Drew Bagnell},
  title        = {A Reduction of Imitation Learning and Structured Prediction to No-Regret
                  Online Learning},
  booktitle    = {{AISTATS} 2011},
}

@inproceedings{DBLP:conf/sbmf/TeSSla,
  author       = {Lukas Convent and
                  Sebastian Hungerecker and
                  Martin Leucker and
                  Torben Scheffel and
                  Malte Schmitz and
                  Daniel Thoma},
  title        = {TeSSLa: Temporal Stream-Based Specification Language},
  booktitle    = {{SBMF} 2018},
}

@inproceedings{DBLP:conf/rv/Striver,
  author       = {Felipe Gorostiaga and
                  C{\'{e}}sar S{\'{a}}nchez},
  editor       = {Christian Colombo and
                  Martin Leucker},
  title        = {Striver: Stream Runtime Verification for Real-Time Event-Streams},
  booktitle    = {{RV} 2018},
}

@incollection{DBLP:series/lncs/BartocciDDFMNS18,
  author       = {Ezio Bartocci and
                  Jyotirmoy V. Deshmukh and
                  Alexandre Donz{\'{e}} and
                  Georgios Fainekos and
                  Oded Maler and
                  Dejan Nickovic and
                  Sriram Sankaranarayanan},
  title        = {Specification-Based Monitoring of Cyber-Physical Systems: {A} Survey
                  on Theory, Tools and Applications},
  booktitle    = {Lectures on Runtime Verification - Introductory and Advanced Topics},
  year         = {2018},
}

@article{DBLP:journals/jlp/LeuckerS09,
  author       = {Martin Leucker and
                  Christian Schallhart},
  title        = {A brief account of runtime verification},
  journal      = {J. Log. Algebraic Methods Program.},
  year         = {2009},
}

@article{DBLP:journals/fmsd/MoosbruggerRS17,
  author       = {Patrick Moosbrugger and
                  Kristin Y. Rozier and
                  Johann Schumann},
  title        = {{R2U2:} monitoring and diagnosis of security threats for unmanned  aerial systems},
  journal      = {Formal Methods Syst. Des.},
  year         = {2017},
}

@inproceedings{DBLP:conf/time/DAngeloSSRFSMM05,
  author       = {Ben D'Angelo and
                  Sriram Sankaranarayanan and
                  C{\'{e}}sar S{\'{a}}nchez and
                  Will Robinson and
                  Bernd Finkbeiner and
                  Henny B. Sipma and
                  Sandeep Mehrotra and
                  Zohar Manna},
  title        = {{LOLA:} Runtime Monitoring of Synchronous Systems},
  booktitle    = {
                  {TIME} 2005},
}

@inproceedings{kopetzki2017methods,
  author       = {Anna{-}Kathrin Kopetzki and
                  Bastian Sch{\"{u}}rmann and
                  Matthias Althoff},
  title        = {Methods for order reduction of zonotopes},
  booktitle    = {CDC 2017},
}

@article{wetzlinger2022adaptive,
  title={Adaptive reachability algorithms for nonlinear systems using abstraction error analysis},
  author={Wetzlinger, Mark and Kulmburg, Adrian and Le Penven, Alexis and Althoff, Matthias},
  journal={NAHS},
  year={2022},
}

@inproceedings{wetzlinger2020adaptive,
  title={Adaptive parameter tuning for reachability analysis of linear systems},
  author={Wetzlinger, Mark and Kochdumper, Niklas and Althoff, Matthias},
  booktitle={CDC 2020},
}

@inproceedings{DBLP:conf/rv/rlola,
  author       = {Bernd Finkbeiner and
                  Martin Fr{\"{a}}nzle and
                  Florian Kohn and
                  Paul Kr{\"{o}}ger},
  title        = {Stream-Based Monitoring Under Measurement Noise},
  booktitle    = {RV 2024},
}

@article{DBLP:journals/automatica/scott,
  author       = {Joseph K. Scott and
                  Davide Martino Raimondo and
                  Giuseppe Roberto Marseglia and
                  Richard D. Braatz},
  title        = {Constrained zonotopes: {A} new tool for set-based estimation and fault
                  detection},
  journal      = {Autom.},
  year         = {2016},
}

@inproceedings{DBLP:conf/fm/copilot,
  author       = {Ivan Perez and
                  Alwyn E. Goodloe and
                  Frank Dedden},
  title        = {Runtime Verification in Real-Time with the Copilot Language: {A} Tutorial},
  booktitle    = {{FM} 2024},
}

@article{DBLP:journals/infsof/UnaaLola,
  author    = {Raik Hipler and
               Hannes Kallwies and
               Martin Leucker and
               Marco Montali and
               C{\'{e}}sar S{\'{a}}nchez and
               Sarah Winkler},
  title     = {Symbolic runtime verification for monitoring under uncertainties and
               assumptions},
  journal   = {Inf. Softw. Technol.},
  year      = {2026},
}

@inproceedings{DBLP:conf/hybrid/Girard05,
  author    = {Antoine Girard},
  title     = {Reachability of Uncertain Linear Systems Using Zonotopes},
  booktitle = {{HSCC} 2005},
  }

@inproceedings{DBLP:conf/eucc/Combastel03,
  author    = {Christophe Combastel},
  title     = {A state bounding observer based on zonotopes},
  booktitle = {ECC 2003},
}

@inproceedings{DBLP:conf/cdc/AlamoB003,
  author    = {Teodoro Alamo and
               Jos{\'{e}} Manuel Bravo and
               Eduardo F. Camacho},
  title     = {Guaranteed state estimation by zonotopes},
  booktitle = {{CDC} 2003},
}

@article{saltik2018outlook,
  title={An outlook on robust model predictive control algorithms: Reflections on performance and computational aspects},
  author={Salt{\i}k, M Bahad{\i}r and {\"O}zkan, Leyla and Ludlage, Jobert HA and Weiland, Siep and Van den Hof, Paul MJ},
  journal={Journal of Process Control},
  year={2018},
}

@phdthesis{lowerre1976harpy,
  title={The HARPY speech recognition system},
  author={Lowerre, Bruce T.},
  year={1976},
  school={Carnegie Mellon University}
}
